\documentclass[runningheads]{llncs}

\usepackage{eccv}

\usepackage{eccvabbrv}
\usepackage{epstopdf}
\usepackage{graphicx}
\usepackage{booktabs}
\usepackage{pifont}
\usepackage{multirow}
\usepackage{wrapfig}
\usepackage[dvipsnames]{xcolor}
\usepackage{booktabs}
\usepackage{makecell}
\usepackage{caption}
\usepackage{graphicx}
\usepackage{float} 
\usepackage{subcaption}
\usepackage[misc]{ifsym}

\usepackage[accsupp]{axessibility}  % Improves PDF readability for those with disabilities.

\usepackage{hyperref}

\usepackage{orcidlink}

\begin{document}

\makeatletter
\def\@fnsymbol#1{%
   \ifcase#1\or
   %\TextOrMath\textasteriskcentered *\or
   \TextOrMath \Letter \dagger\or
   \TextOrMath \textdagger \ddagger\else
   \@ctrerr \fi
}
\makeatother

\title{DailyDVS-200: A Comprehensive Benchmark Dataset for Event-Based Action Recognition} 

\titlerunning{DailyDVS-200 for Event-Based Action Recognition}

\author{Qi Wang\inst{1*}\orcidlink{0009-0004-4576-4458}\and
Zhou Xu\inst{1*}\orcidlink{0009-0004-5499-0537} \and
Yuming Lin\inst{1}\orcidlink{0009-0009-7439-7462} \and 
Jingtao Ye\inst{1}\orcidlink{0009-0002-6636-5072} \and
Hongsheng Li\inst{1}\orcidlink{0000-0002-9929-4023} \and
Guangming Zhu\inst{1}\orcidlink{0000-0003-3214-4095} \and
Syed Afaq Ali Shah\inst{2}\orcidlink{0000-0003-2181-8445} \and
Mohammed Bennamoun\inst{3}\orcidlink{0000-0002-6603-3257} \and
Liang Zhang\inst{1}\thanks{ Corresponding author. * Equal contribution.}\orcidlink{0000-0003-4331-5830}
}

\authorrunning{Q.~Wang et al.}

% \institute{Xidian University, School of Computer Science and Technology, China \email{\{qiwang0720, xuzhou0112, linyuming, hsli\}@stu.xidian.edu.cn}\\
% \email{yejingtao030920@qq.com, \{gmzhu, liangzhang\}@xidian.edu.cn} \and
% Edith Cowan University \email{afaq.shah@ecu.edu.au} \and
% University of Western Australia \email{mohammed.bennamoun@uwa.edu.au}
% }
\institute{Xidian University, School of Computer Science and Technology, China \email{\{qiwang0720, xuzhou0112\}@stu.xidian.edu.cn, liangzhang@xidian.edu.cn} \and
Edith Cowan University \email{afaq.shah@ecu.edu.au} \and
University of Western Australia \email{mohammed.bennamoun@uwa.edu.au}
}

\maketitle

\begin{abstract}
  Neuromorphic sensors, specifically event cameras, revolutionize visual data acquisition by capturing pixel intensity changes with exceptional dynamic range, minimal latency, and energy efficiency, setting them apart from conventional frame-based cameras. The distinctive capabilities of event cameras have ignited significant interest in the domain of event-based action recognition, recognizing their vast potential for advancement. However, the development in this field is currently slowed by the lack of comprehensive, large-scale datasets, which are critical for developing robust recognition frameworks. To bridge this gap, we introduces \textbf{\emph{DailyDVS-200}}, a meticulously curated benchmark dataset tailored for the event-based action recognition community. DailyDVS-200 is extensive, covering 200 action categories across real-world scenarios, recorded by 47 participants, and comprises more than 22,000 event sequences. This dataset is designed to reflect a broad spectrum of action types, scene complexities, and data acquisition diversity. Each sequence in the dataset is annotated with 14 attributes, ensuring a detailed characterization of the recorded actions. Moreover, DailyDVS-200 is structured to facilitate a wide range of research paths, offering a solid foundation for both validating existing approaches and inspiring novel methodologies. By setting a new benchmark in the field, we challenge the current limitations of neuromorphic data processing and invite a surge of new approaches in event-based action recognition techniques, which paves the way for future explorations in neuromorphic computing and beyond. The dataset and source code are available at \url{https://github.com/QiWang233/DailyDVS-200}. 
\end{abstract}

\section{Introduction}
\label{sec:intro}

Action recognition is of significant importance across various domains, spanning from intelligent surveillance to video understanding. Current action recognition algorithms, leveraging large-scale frame-based benchmark datasets, have demonstrated remarkable performance. However, several challenges arise concerning the storage, transmission, and analysis of frame data. Redundant frames in video data contribute to unnecessary storage requirements and escalate power consumption burdens on devices. Moreover, frame-based cameras suffer from low frame rates and limited dynamic ranges, impeding their capability to effectively capture fast-moving objects and operate optimally under challenging lighting conditions like back-lighting and low light. So, the exploration and implementation of innovative solutions become imperative to address these limitations.

\begin{figure}[tb]
  \centering
  \includegraphics[height=7.3cm]{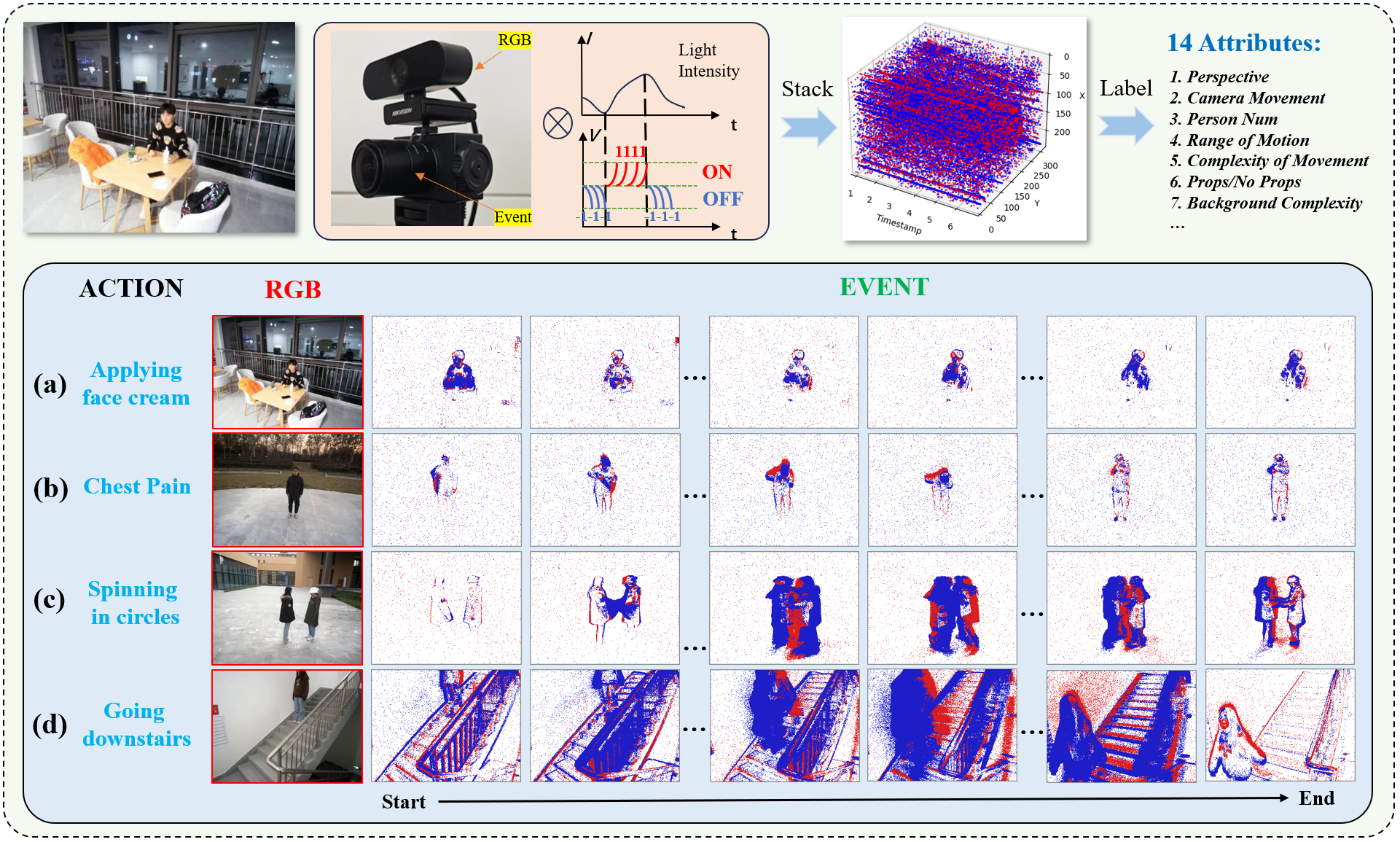}
  \caption{The flow of our data acquisition process. We use both an RGB camera (above) and a DVS camera (below). Upon completion of the recording, the DVS camera generates event flow data, while the RGB camera captures the synchronized video stream. Subsequently, the data is processed to remove noise, and each sample is categorized based on its motion characteristics.
  }
  \label{fig:heading}
  % \vspace{-.4cm} 1
\end{figure}

\begin{figure}[tb]

	\centering
	\begin{subfigure}{0.50\linewidth}
		\centering
		\includegraphics[width=1\linewidth]{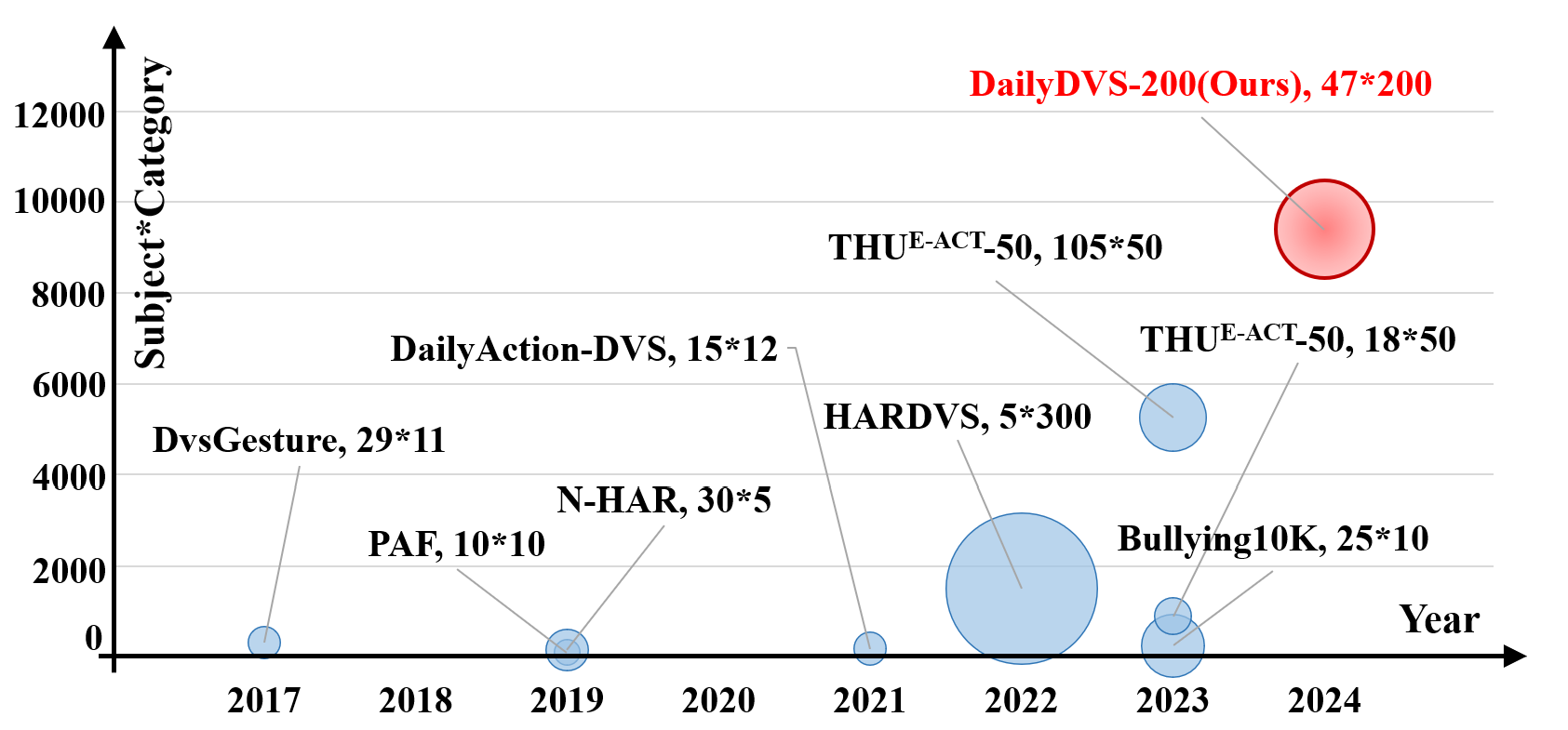}
		% \caption{}
		\label{chutian1}%文中引用该图片代号
	\end{subfigure}
	% \hspace{0.1mm}
	\begin{subfigure}{0.45\linewidth}
		\centering
		\includegraphics[width=1\linewidth]{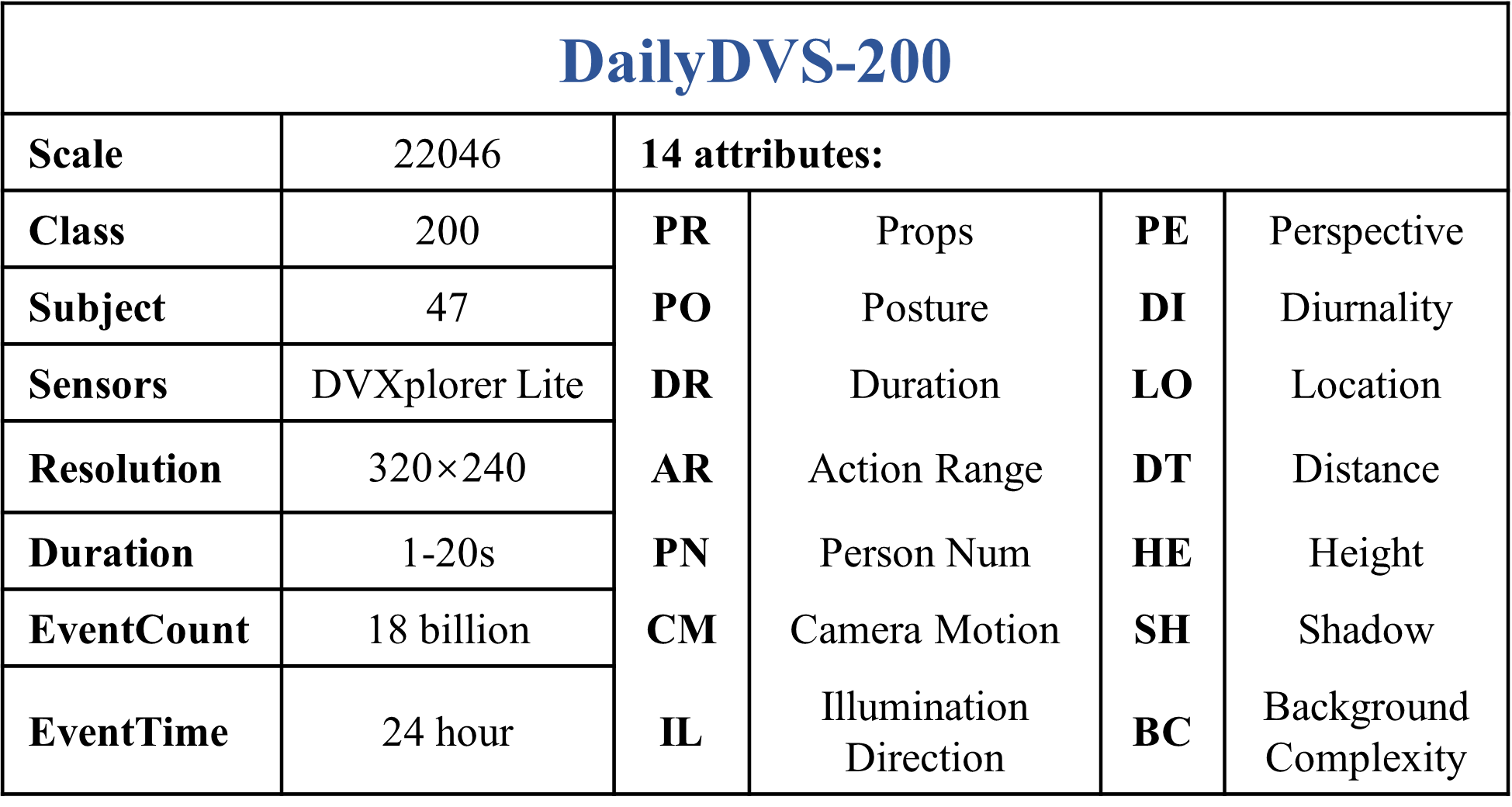}
		% \caption{}
		\label{chutian2}%文中引用该图片代号
	\end{subfigure}
        \caption{(Left) A comparison between existing datasets and our proposed DailyDVS-200 dataset for event-based action classification. (Right) Summary of Characteristics of DailyDVS-200. The nomenclature is \textbf{PR}: Props, \textbf{PO}: Posture, \textbf{DR}: Duration, \textbf{AR}: Action Range, \textbf{PN}: Person Num, \textbf{CM}: Camera Motion, \textbf{IL}: Illumination Direction, \textbf{PE}: Perspective, \textbf{DI}: Diurnality, \textbf{LO}: Location, \textbf{DT}: Distance, \textbf{HE}: Height, \textbf{SH}: Shadow, \textbf{BC}: Background Complexity.}
        \label{fig:group}
        % \vspace{-.4cm} 1
\end{figure}

The development of dynamic vision sensors (also known as event cameras), such as DAVIS\cite{6889103}, CeleX\cite{chen2019live}, ATIS\cite{posch2010qvga}, and PROPHESEE\footnote[1]{\url{https://www.prophesee.ai}}, has introduced a new paradigm in visual perception. Unlike traditional cameras that capture images at fixed exposure rates, event cameras asynchronously record points in the scene where pixel brightness changes exceed a certain threshold, and output event in the form of tuples ($t, x, y, p$), where $t$ represents the timestamp, ($x, y$) represents the two-dimensional coordinates, and $p$ represents the polarity. Additionally, stacking events over a period of time can be viewed as a discrete 3D sequence along the time axis, as illustrated in the first row of \cref{fig:heading}. Since event cameras only capture parts of the scene where brightness changes occur, they significantly reduce redundant information and lower the storage and computational load on devices\cite{li2022retinomorphic, wang2108reliable, zhu2021neuspike, zhu2022event}. Moreover, event streams emphasize the approximate outline of objects without recording specific color and texture features like traditional cameras, thus greatly protecting user privacy and eliminating concerns about privacy leakage during the use of related devices in daily life. Event cameras also exhibit high frame rates and a wide dynamic range, leading to densely packed event streams over time with minimal motion blur, enabling them to effectively capture fast-moving human actions. These attributes render them highly suitable for addressing the current challenges in action recognition.

In an ideal scenario, action recognition primarily involves systems analyzing motion characteristics of the human body trunk to predict and classify actions. However, in real-life situations, recognition systems not only capture human motion information but also record redundant information such as environmental backgrounds, as shown in the second row of \cref{fig:heading}, action (a)(b)(c) have no significant background information, while (d) prominently features staircase details. Thus, key human motion information is often disrupted by various factors such as camera motion, light intensity, scene complexity, \etc. These factors affect the integrity of the data and increase the challenges of correct recognition. 

\begin{table}[tb]
\caption{Comparison of event datasets for action recognition. Sub, MA, AA and DR denotes Subject, Multi-Attribute, Attribute Annotation and Duration of the action, respectively. Note that we only report these groups of real DVS datasets.
}
\label{tab:dataset}
\tabcolsep=4pt
\centering
\resizebox{\textwidth}{!}{
\begin{tabular}{lcccccccccc}
\toprule
\textbf{Dataset}       & \textbf{Year} & \textbf{Sensors} & \textbf{Object} &\textbf{Scale} & \textbf{Class}  & \textbf{Sub}  & \textbf{Real} & \textbf{MA} & \textbf{AA} & \textbf{DR}     \\ \midrule
ASLAN-DVS \cite{bi2020graph}  & 2011          & DAVIS240c       & Action & 3,697           & 432            & -                          & \ding{55}             & -           & -  & -      \\
MNISTDVS\cite{serrano2015poker}      & 2013          & DAVIS128      & Image & 30,000          & 10             & -                             & \ding{55}             & -           & -  & -      \\
N-Caltech101\cite{orchard2015converting}   & 2015          & ATIS      & Image       & 8,709           & 101            & -                             & \ding{55}             & -           & - & 0.3s      \\
N-MNIST\cite{orchard2015converting}        & 2015          & ATIS     & Image        & 70,000          & 10             & -                              & \ding{55}             & -           & -  & 0.3s      \\
CIFAR10-DVS\cite{li2017cifar10}    & 2017          & DAVIS128    & Image     & 10,000          & 10             & -                            & \ding{55}             & -           & -  & 1.2s      \\
HMDB-DVS\cite{bi2020graph, kuehne2011hmdb}     & 2019          & DAVIS240c    & Action    & 6,766           & 51             & -                           & \ding{55}             & -           & -  & 19s      \\
UCF-DVS\cite{bi2020graph, soomro2012ucf101}      & 2019          & DAVIS240c    & Action    & 13,320          & 101            & -                            & \ding{55}             & -           & -  & 25s     \\
N-ImageNet\cite{kim2021n}     & 2021          & Samsung-Gen3  & Image   & 1,781,167        & 1,000           & -                            & \ding{55}             & -           & -  & -      \\
ES-lmageNet\cite{lin2021imagenet}    & 2021          & -      & Image          & 1,306,916        & 1,000           & -                         & \ding{55}             & -           & -  & -      \\ \midrule
DvsGesture\cite{amir2017low}     & 2017          & DAVIS128    & Action     & 1,342           & 11             & 29                           & \ding{52}             & \ding{55}           & \ding{55}  & 6s      \\
N-CARS\cite{sironi2018hats}         & 2018          & ATIS   & Car          & 24,029          & 2              & -                           & \ding{52}             & \ding{55}           & \ding{55}  & 0.1s     \\
ASL-DVS\cite{bi2020graph}         & 2019          & DAVIS240   & Hand      & 100,800         & 24             & 5                           & \ding{52}             & \ding{55}           & \ding{55}  & 0.1s   \\
PAF\cite{miao2019neuromorphic}            & 2019          & DAVIS346  & Action       & 450            & 10             & 10                          & \ding{52}             & \ding{55}           & \ding{55}  & 5s     \\
DailyAction\cite{liu2021event}    & 2021          & DAVIS346   & Action      & 1,440           & 12             & 15                         & \ding{52}             & \ding{55}           & \ding{55}  & 5s     \\
HARDVS \cite{wang2022hardvs}    & 2022          & DAVIS346   & Action      & 107,646         & 300            & 5                            & \ding{52}             & \ding{52}           & \ding{55}  & 5s     \\
\( \mathrm{THU}^\mathrm{E\text{-}ACT} \)-50\cite{gao2023action}   & 2023          & CeleX-V   & Action       & 10,500          & 50             & 105                         & \ding{52}             & \ding{55}           & \ding{55}  & 2-5s  \\
\( \mathrm{THU}^\mathrm{E\text{-}-ACT} \)-50-CHL\cite{gao2023action}   & 2023          & DAVIS346   & Action       & 2,330          & 50             & 18                         & \ding{52}             & \ding{55}           & \ding{55}  & 2-5s  \\
Bullying10K\cite{dong2024bullying10k} & 2023          & DAVIS346  & Action       & 10,000          & 10             & 25                          & \ding{52}             & \ding{55}           & \ding{55}  & 2-20s  \\ \midrule
DailyDVS-200 \textbf{(Ours)}          & 2024          & DVXplorer Lite & Action  & 22,046          & 200            & 47                          & \ding{52}             & \ding{52}           & \ding{52}  & 1-20s \\ \bottomrule
\end{tabular}
}
% \vspace{-0.4cm} 1
\end{table}

Although some benchmark datasets have been proposed, many of them are synthetic datasets captured by pointing event cameras at screens displaying RGB images \cite{orchard2015converting}, or by converting commonly used RGB action recognition datasets into simulated event streams \cite{bi2020graph, kuehne2011hmdb, soomro2012ucf101, kliper2011action}. However, simulated datasets often result in information loss due to their multi-stage nature, making it difficult to achieve the effectiveness of real event datasets in practice. While several large-scale real-world benchmark event datasets have been proposed recently \cite{amir2017low, bi2020graph, sironi2018hats, miao2019neuromorphic, liu2021event, wang2022hardvs, li2022n, gao2023action, dong2024bullying10k, duarte2024event}, such as DvsGesture \cite{amir2017low}, PAF \cite{miao2019neuromorphic}, and Hardvs \cite{wang2022hardvs}, they are limited by small scale, few categories, and limited diversity of individuals. Additionally, \( \mathrm{THU}^\mathrm{E\text{-}ACT} \)-50 \cite{gao2023action} consists only of simple actions in fixed scenes, and although \( \mathrm{THU}^\mathrm{E\text{-}ACT} \)-50-CHL \cite{gao2023action} introduces challenging actions, it is still limited by the number and diversity of actions. Bullying10K \cite{dong2024bullying10k} focuses exclusively on 10 types of bullying actions, making it difficult to demonstrate the superiority of models. Therefore, there is an urgent need for a comprehensive dataset that takes into account various factors to address these issues.

To bridge these aforementioned gaps, we present a novel and comprehensive dataset, termed DailyDVS-200. Our proposed dataset consists of 200 distinct action categories, collected from 47 subjects. As shown in \cref{fig:group} (left), the dataset we provide has the richest number of subject-category combinations. The detailed comparison with existing benchmark datasets can be found in \Cref{tab:dataset}. Our dataset systematically incorporates a wide range of real-life scenarios, including variations in \emph{viewpoint}, \emph{diurnal shifts}, \emph{indoor and outdoor settings}, \emph{lighting conditions}, \emph{camera movements}, \emph{actor counts}, \emph{action duration}, \emph{shadow effects}, \emph{shooting elevations}, \emph{distances}, \emph{prop presence}, \emph{action scopes}, \emph{background complexities}, and \emph{pose diversity}. Moreover, each data is annotated based on these attributes, more details can be found in \cref{fig:group} (right) and \cref{fig:data_static} (a).

Leveraging our newly introduced DailyDVS-200 dataset, we conduct a comprehensive evaluation of four distinct types of action recognition deep models across over ten diverse frameworks. Our findings reveal that the current state-of-the-art event-based action recognition networks continue to underperform when compared to traditional action recognition frameworks. \emph{Additionally, we divided the dataset into 14 groups based on the annotation of attributes}. Through diverse group evaluations, we showed significant variations in the model's recognition performance under different action conditions. For instance, under the Swin-T\cite{liu2022video} test, recognition performance significantly differs between dynamic and static camera conditions (27.84\% vs. 58.05\%), as well as across diverse action ranges (Full-body: 59.24\% vs. Limbs: 43.15\% vs. Micro: 34.59\%). Additionally, we undertake parallel tests under identical conditions using existing large-scale event datasets to authenticate the challenging nature of our dataset. In summary, our contributions are mainly reflected in the following aspects:
\begin{itemize}
\setlength{\topsep}{2pt}
\setlength{\parsep}{2pt}
\setlength{\parskip}{2pt}
\item We propose a large-scale neuromorphic dataset for action recognition, named DailyDVS-200. It consists of over 22k samples collected from 47 subjects, spanning 200 categories, and comprehensively reflects real-world challenges, with corresponding attribute annotations provided for each data point. \emph{To the best of our knowledge, this is the first large-scale real-world neuromorphic dataset for action recognition with label provided}.
\item To objectively evaluate the performance of different methods, we establish benchmarks for various types of recognition models on our dataset. This provides a wide baseline for future comparisons on the DailyDVS-200 dataset.
\item We conducted group evaluation and analysis based on the annotation of attributes. This not only validates the impact of different attribute on event-based action recognition models but also introduces new research content for neuromorphic datasets, thereby driving advancements in the field.
\end{itemize}

\section{Related Work}
\subsection{Event-based Dataset}
\subsubsection{Synthetic Datasets}
Currently, there are various publicly available event datasets, but early DVS datasets mainly originated from existing image classification datasets\cite{li2017cifar10, orchard2015converting}. They primarily captured changes in brightness using DVS cameras and the relative motion of scenes, which posed significant challenges. Subsequently, Bi \etal\cite{bi2020graph} expanded the number of event datasets by converting publicly available frame-based action datasets into simulated event streams, including HMDB-DVS\cite{kuehne2011hmdb}, ASLAN-DVS\cite{kliper2011action}, and UCF-DVS\cite{soomro2012ucf101}. However, these synthetic datasets somewhat suppressed the characteristics of event cameras, such as high dynamic range and high frame rate, by initially capturing images with regular cameras and then manually converting them into event streams.

\subsubsection{Real-World Dataset for Event-Based Action Recognition}
In recent years, small-sized real-world event-based action datasets have emerged\cite{amir2017low, miao2019neuromorphic, liu2021event}, but they have limitations. DvsGesture\cite{amir2017low} has only 11 action categories with data from 29 individuals, PAF\cite{miao2019neuromorphic} includes only 10 actions and 450 records, and DailyAction\cite{liu2021event} comprises 12 actions with 1,440 records at low resolution (128×128). Larger datasets like \( \mathrm{THU}^\mathrm{E\text{-}ACT} \)-50\cite{gao2023action} offer more samples but have limitations in term of consideration of challenging scenarios. Hardvs\cite{wang2022hardvs} is currently the largest dataset in terms of data volume, but it also has the smallest number of participants, with only 5 individuals, which presents significant limitations. Bullying10k\cite{dong2024bullying10k} focuses on a single bullying-related scene and lacks diversity for comprehensive action recognition research. These datasets, despite their merits, do not sufficiently advance event-based action recognition models or fully leverage the potential of event cameras.

\subsection{Event Representation}
Due to the asynchronous and discrete nature of event streams, they cannot be directly used for training. Therefore, it is necessary to convert them into suitable alternative representations for different action recognition methods. Frame-based representations of event streams are currently widely used because they allow simple generation of event frames by aggregating events within specific time windows\cite{zhu2019unsupervised, moeys2016steering, lagorce2016hots, sironi2018hats}. However, such designs primarily extract spatio-temporal information and do not fully exploit the time and polarity information contained in events. With the widespread adoption of Transformer networks in various event-based tasks, there has been a recent focus on converting events into suitable token forms\cite{sabater2022event, de2023eventtransact, peng2023get}. For instance, Peng \etal \cite{peng2023get} proposed a novel Group Token that reorganizes asynchronous events based on their timestamps and polarities, effectively leveraging the characteristics of events. Some researchers are also exploring new neural network models, such as Spiking Neural Networks (SNNs)\cite{neftci2019surrogate, che2022differentiable, zeng2023braincog}, which simulate the pulse transmission of neurons to achieve information processing, making them more akin to the working mechanism of biological neural systems. Additionally, there are other representation methods, such as learning-based representations\cite{cannici2020differentiable, messikommer2020event, gehrig2019end}, graph-based representations\cite{li2021graph, schaefer2022aegnn} and TORE \cite{baldwin2022time}. However, it remains unclear which representation method is most suitable for event streams at present.

\subsection{Action Recognition}
Historically, in frame-based action recognition, there have typically been two key steps: action representation\cite{laptev2005space, wang2016robust, scovanner20073, morency2007latent} and action classification\cite{kong2022human, liu2011recognizing, shi2011human}. In recent years, deep learning techniques have integrated these two steps into an end-to-end learning framework, significantly improving action classification performance. To leverage information from all frames and model the inter-frame information correlation, Tran \etal \cite{tran2015learning} proposed 3DCNN to learn features in both spatial and temporal domains, but with high computational costs. Carreira and Zisserman \cite{carreira2017quo} introduced I3D, which adapts well established image classification architectures, making training easier. Feichtenhofer \etal \cite{feichtenhofer2019slowfast} proposed an efficient network, SlowFast, with both slow and fast pathways that can adapt to different scenarios by adjusting channel capacities, greatly enhancing overall efficiency. Additionally, various 3DCNN variants \cite{feichtenhofer2020x3d, zhu2020a3d, tran2018closer} have been proposed, further improving recognition efficiency. With the introduction of ViT\cite{dosovitskiy2020image}, self-attention mechanisms\cite{vaswani2017attention, devlin2018bert} have been applied to action recognition\cite{liu2022video, bertasius2021space}, which has been shown to achieve good performance. Spiking neural networks have also been used for action recognition. However, due to the non-differentiability of discrete pulse signals, training SNNs poses challenges. Several effective training methods have been proposed to address this challenge\cite{neftci2019surrogate, che2022differentiable, zeng2023braincog, yao2024spike, zhou2022spikformer}, but their effectiveness remains to be further investigated.

\section{DailyDVS-200 Dataset}
In this section, we provide a detailed overview of the data acquisition, annotation methodology, and analysis of action categories for our DailyDVS-200 dataset.

\subsection{Data Structure}
\subsubsection{Data Modalities}
To meet the computation requirements of real life applications, we used a DVXplorer Lite sensor paired with an RGB camera to collect our dataset, details are shown in \cref{fig:heading}. We employed custom-built software to capture event data and save it as standard \emph{aedat4} files, which include the event stream, IMU stream, and trigger stream. The spatial resolution of the event camera is 320×240, and the RGB camera is synchronized to ensure the quality of the capture and assist with subsequent annotation tasks.
% \vspace{-0.2cm} 1
\subsubsection{Subjects}
To ensure the diversity and authenticity of the data, we selected 47 different subjects (26 males and 21 females) from hundreds of participants based on factors such as gender, physique, and height to participate in our dataset collection work. Each participant was assigned a unique ID number.
% \vspace{-0.2cm} 1
\subsubsection{Action Classes}
\begin{figure}[tb]
  \centering
  \includegraphics[height=7.2cm]{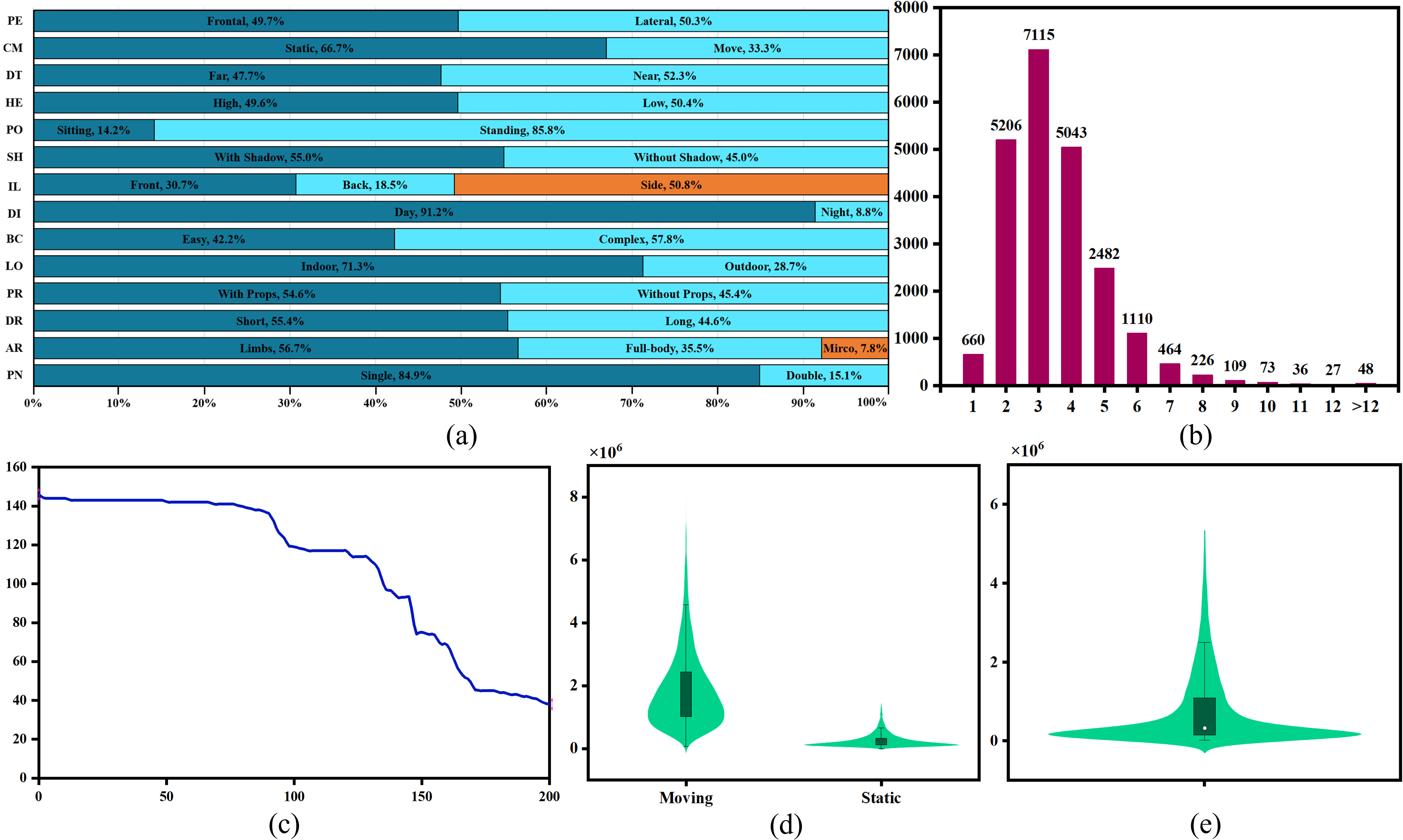}
  \caption{Statistical data and analysis of DailyDVS-200. (a) Data proportions for the 14 attributes. (b) Distribution of data volumes for different time duration in seconds. (c) Number of images per class. (d) Distribution of Event Count compared between the moving and static. (e) Distribution of Event Count for all categories.
  }
  \label{fig:data_static}
  % \vspace{-.4cm} 1
\end{figure}

To enhance the practical applicability of our DailyDVS-200 dataset, we meticulously curated and supplemented 200 daily action categories from commonly used video-based public datasets \cite{liu2019ntu, kay2017kinetics, soomro2012ucf101, caba2015activitynet}, a selection process that resulted in a dataset closely mirroring real-life scenarios. Finally, our dataset comprises 22,046 records, more detail about our DailyDVS-200 dataset can be seen in \cref{fig:group} (right) and \cref{fig:data_static}.
 % Finally, as shown in \cref{fig:group} (right), our dataset comprises 22,046 records, with each record covering a time span of 1-20 seconds, amounting to a total duration of 24 hours. The distribution of different actions and the event distribution of the data can be seen in \cref{fig:data_static} (c, e).
 
The DailyDVS-200 dataset includes the following scenes: (1) \textbf{Household Activities}: \emph{Sweeping}, \emph{mopping}, \emph{combing hair}, \emph{washing towels}, \emph{folding clothes}, \emph{brushing teeth}, \emph{washing face}, \emph{cutting nails}, \etc.
(2) \textbf{Office Tasks}: \emph{Nodding}, \emph{clapping}, \emph{sitting down}, \emph{standing up}, \emph{typing on a keyboard}, \emph{moving a mouse}, \emph{opening drawers}, \emph{opening a laptop}, \etc.
(3) \textbf{Sports and Physical Activities}: \emph{Jumping in place}, \emph{running}, \emph{dribbling basketball}, \emph{long jump}, \emph{skipping rope}, \emph{push-ups}, \emph{kicking a ball}, \emph{swinging a badminton racket}, \etc.
(4) \textbf{Health-related Activities}: \emph{Headache}, \emph{chest pain}, \emph{back pain}, \emph{vomiting}, \emph{leg massage}, \etc.
(5) \textbf{Interactions}: \emph{Handshaking}, \emph{toasting}, \emph{hugging}, \emph{high-fiving}, \emph{arm wrestling}, \emph{fist bumping}, \etc.
(6) \textbf{Bullying and Violence}: \emph{Fighting}, \emph{hitting}, \emph{kicking}, \emph{pushing}, \emph{using objects to attack}, \etc.
(7)\textbf{ Transportation-related Activities}: \emph{Riding a bicycle}, \emph{riding an electric scooter}, \emph{walking with a backpack}, \etc. This diverse range of action categories ensures the relevance of the proposed dataset to various real-world applications and provides a comprehensive basis for event-based action recognition research.

% \begin{figure}[tb]
%   \centering
%   \includegraphics[height=7cm]{preview.png}
%   \caption{A preview of DailyDVS-200 dataset.
%   }
%   \label{fig:preview}
% \end{figure}

\begin{figure}[tb]
  \centering
  \includegraphics[height=6.9cm]{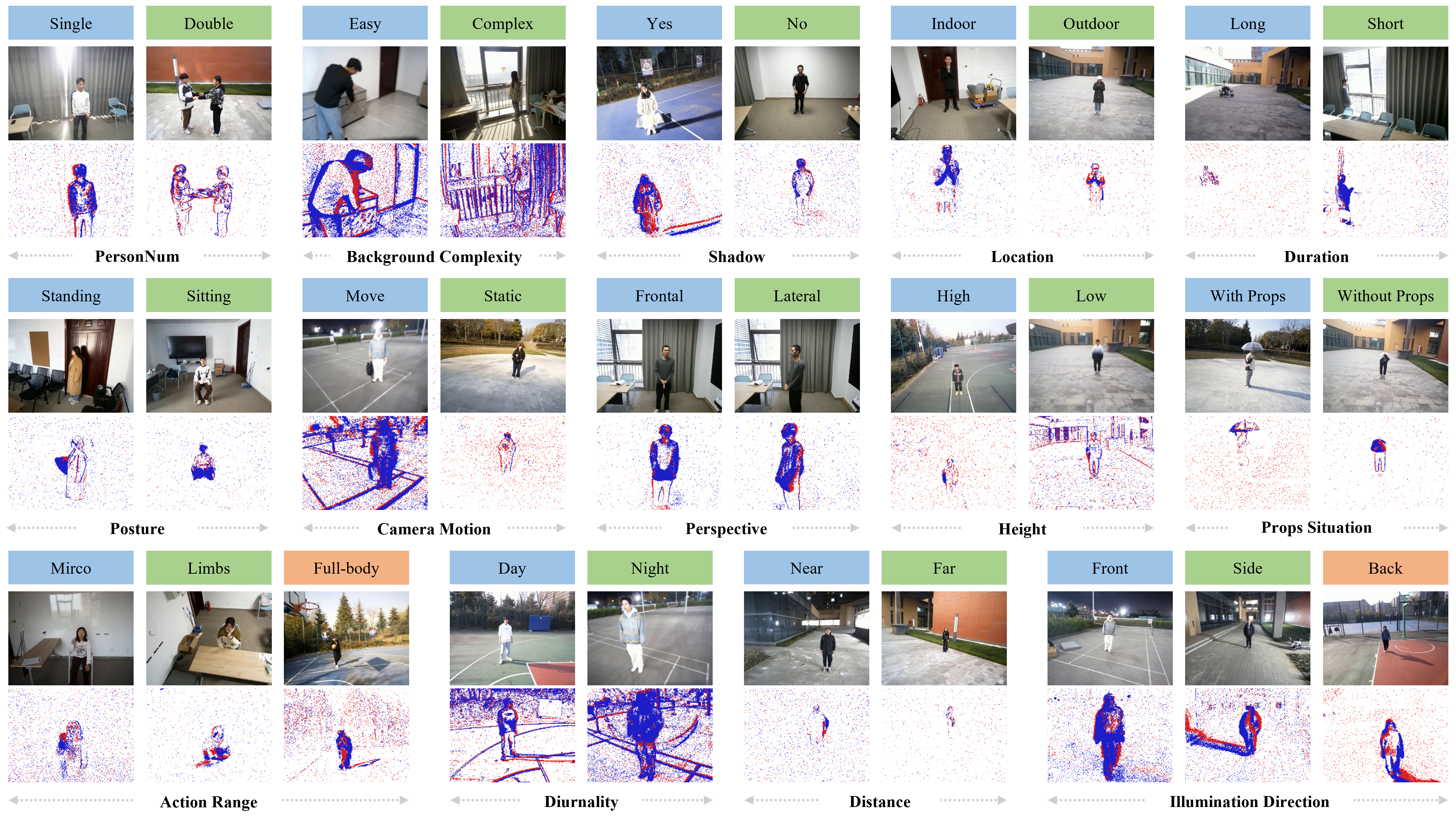}
  \caption{A preview of our proposed DailyDVS-200 dataset and examples of our attribute annotations.
  }
  \label{fig:preview}
% \vspace{-.4cm} 1
\end{figure}

% \begin{figure}[tb]
% 	\centering
% 	\begin{subfigure}{0.49\linewidth}
% 		\centering
% 		\includegraphics[width=4.8cm]{static_move2.png}
% 		% \caption{}
% 		\label{chutian1}%文中引用该图片代号
% 	\end{subfigure}
%  	% \hspace{1mm}
% 	\centering
% 	\begin{subfigure}{0.49\linewidth}
% 		\centering
% 		\includegraphics[width=4.8cm]{time2.png}
% 		% \caption{}
% 		\label{chutian2}%文中引用该图片代号
% 	\end{subfigure}
%         \caption{(Left) Distribution of Event Count in DVS Camera during Movement and Static States. (Right) Distribution of Action Duration in DailyDVS-200.}
%         \label{fig:move_static}
% \end{figure}

The dataset consists of a wide range of daily actions, considering various action characteristics:
(1) \textbf{Fine-grained Micro, Limb, and Whole-body Movements}: Examples include finger movements such as \emph{writing} and \emph{trimming nails}, limb movements such as \emph{rotating one arm} and \emph{checking the time} as well as whole-body movements like \emph{Tai Chi} and \emph{walking with an umbrella}.
(2) \textbf{Short-duration and Long-duration Actions}: It covers simple actions like \emph{clapping hands} and \emph{nodding} compared to longer actions such as \emph{putting on shoes} and \emph{dressing}. The time distribution of our dataset is illustrated in \cref{fig:data_static} (b). It can be observed that we have a diverse range of action duration.
(3) \textbf{Actions with and without Props Interaction}: It includes actions involving interactions with objects such as \emph{playing volleyball} and \emph{playing table tennis}, as well as actions without object interaction like \emph{going upstairs} and \emph{going downstairs}.

% \begin{figure}[tb]
%   \centering
%   \includegraphics[height=4cm]{attr.png}
%   \caption{Number of attribute labels.
%   }
%   \label{attr:preview}
% \end{figure}

In addition, as shown in \cref{fig:preview}, we pay close attention to the diverse performances of the same action under different settings: (1) \textbf{Different Perspectives}: We collect different views of the same action to simulate real-world perspectives, including front view, side view, top-down view, and bottom-up view.
(2) \textbf{Different Distances}: Due to the limitations of the frame, the performance of the same action at different distances often varies significantly. Therefore, we capture actions from distances up to approximately 30 meters.
(3) \textbf{Lighting Conditions}:
    i) Light Direction: Different light directions directly affect the performance of actions in the frame due to the event camera's sensitivity to light. Therefore, we collect data under front light, side light, and back light conditions.
    ii) Day and Night Conditions: Lighting conditions vary between day and night, which directly affect the performance of actions. Additionally, due to the high frame rate of event cameras, the flickering effect of low-frequency lights also has a significant impact.
(4) \textbf{Different Camera Movements}: Event cameras focus only on the changes in light in the frame, making it easier to identify actions under fixed camera conditions. However, such data biases severely limit the application scenarios of event cameras. Therefore, we capture action data under certain camera movements, including varying degrees of background complexity. 

In addition to these factors, which we believe have a significant impact on event data, we also consider other possible scenarios that may occur in real-world settings, including indoor and outdoor settings, sitting and standing postures for the same action, and the influence of shadows.

% \vspace{-0.4cm} 1
\subsubsection{Collection Setups}
To replicate various scenarios encountered in daily life realistically, we set different data acquisition conditions for each participant and selected diverse locations such as different rooms, corridors, open squares, roads, and playgrounds. This ensures the richness and diversity of the scenes. Additionally, our data collection sessions are divided into three time periods: \emph{morning}, \emph{afternoon}, and \emph{evening}, which closely reflect different conditions in real life. We have a variety of scenes to ensure the diversity of our dataset. During data collection, we select actions based on the current scene, and each action is captured three times: \emph{once from a front view}, \emph{once from a side view}, and \emph{once from any view under camera movement}. Camera movement is achieved using a professional photography stand equipped with a mobile chassis. For each participant, we also vary the camera's height (up to 3 meters) and data acquisition distance (up to 30 meters) to obtain multiple shooting perspectives.
% \vspace{-0.6cm} 1
\subsubsection{Data Annotation}
After completing data acquisition for the action recognition task, we performed classification and labeling for each sample to ensure accurate identification of the represented actions. For each behavioral data, we further organized and divided them based on the scene, participant number, and camera movement status. Each category was named according to the participant's number, action name, scene number, and camera status to ensure the clarity and accuracy of the dataset's organizational structure. In \cref{fig:preview}, which provides examples of several groups labels, it can be observed that each attributes displays distinct differences, for example, \emph{Illumination Direction} in the bottom-right corner. We can distinguish the approximate direction of light based on the shadows in the scene. Moreover, we conducted detailed attributes annotations for each data point based on its characteristics. We believe that the granular annotations provided for event-based action recognition can facilitate a deeper understanding of event data.
% \vspace{-0.6cm}
\subsection{Benchmark Evaluations}
To conduct standardized evaluations of the models tested on our benchmark dataset, we have defined precise criteria for two types of action classification assessments. The accuracy is reported as a percentage for each criterion. We utilize 4 NVIDIA GeForce RTX 4090 GPUs for all of the training and testing.
% \vspace{-0.2cm}
\subsubsection{Cross-Subject Evaluation}
For the cross-subject evaluation, the 47 participants were divided into three groups: training set, validation set, and test set. The validation set consists of 8 individuals with the following IDs: 3, 4, 5, 24, 27, 31, 41, 43. The test set comprises 9 individuals with IDs: 4, 7, 10, 11, 16, 33, 37, 42, 45. The remaining 31 individuals form the training set.
% \vspace{-0.2cm}
\subsubsection{Multi-Group Evaluation}
For multi-group evaluation, the model training settings are the same as those used in the Cross-Subject Evaluation. We first conduct training from scratch across participants, and then perform separate testing on the test set with different settings for various groups.

\section{Experimental Results}
\label{sec:blind}

In this section, we employ various methods to conduct cross-subject and multi-group evaluations on our dataset. Firstly, we test our dataset using multiple approaches, including frame-based, learnable-based, token-based, and spike-based methods. Secondly, within the frame-based approach, we explore the impact of frame gap and time step for generating event frames on the experiment results of our dataset. Additionally, we demonstrate the performance of multiple groups in various event-based action recognition models. Finally, we validate our dataset's complexity using standard experimental setups from a large-scale action recognition dataset, with details and findings in the supplementary material.

\subsection{Evaluations of Action Recognition models}
\subsubsection{Evaluation of different methods}
Our dataset was tested on 12 different action recognition algorithms, namely, C3D\cite{tran2015learning}, I3D \cite{carreira2017quo}, R2Plus1D \cite{tran2018closer}, Slowfast \cite{feichtenhofer2019slowfast}, TSM \cite{lin2019tsm}, EST \cite{gehrig2019end}, TimeSformer \cite{bertasius2021space}, Swin-T \cite{liu2022video}, ESTF \cite{wang2022hardvs}, GET \cite{peng2023get}, Spikformer\cite{zhou2022spikformer}, and SDT\cite{yao2024spike}. This section employed cross-subject evaluation criteria, and our experimental results are reported in \Cref{tab:models}. As can be noted, among the frame-based methods (generated at intervals of 500ms), SlowFast\cite{feichtenhofer2019slowfast} achieved the highest top-1 accuracy of 41.49\%, while the TSM\cite{lin2019tsm} achieved the highest top-5 accuracy of 71.46\%. As for the token-based methods, Swin-T\cite{liu2022video} achieved the highest accuracy in both top-1 and top-5, with 48.06\% and 74.47\% respectively, and also achieved the highest accuracy among all methods. In the spike-based methods, Spikformer achieved the highest top-1 and top-5 accuracy of 36.94\% and 62.37\%, respectively.

\begin{table}[tb]
\caption{Evaluation of different methods on our dataset. Best models results with different input types are \textbf{highlighted}.
    }
\label{tab:models}
\centering
\tabcolsep=10pt
\resizebox{1\textwidth}{!}{
\begin{tabular}{@{}llcccc@{}}
\toprule
\textbf{Methods} & \textbf{Year} & \textbf{Input Type} & \textbf{Backbone} &\textbf{ top-1 acc.(\%)} & \textbf{top-5 acc.(\%)} \\ \midrule
C3D\cite{tran2015learning}               & 2015          & Frame               & 3D CNN            & 21.99                   & 45.81                   \\
I3D \cite{carreira2017quo}               & 2017          & Frame               & ResNet50          & 32.30                    & 59.05                   \\
R2Plus1D\cite{tran2018closer}            & 2018          & Frame               & ResNet34          & 36.06                   & 63.67                   \\
SlowFast\cite{feichtenhofer2019slowfast} & 2019          & Frame               & ResNet50          & \textbf{41.49}                   & 68.19                   \\
TSM \cite{lin2019tsm}                    & 2019          & Frame               & ResNet50          & 40.87                   & \textbf{71.46}                   \\ 
EST \cite{gehrig2019end}                 & 2019          & Learnable           & ResNet34          & \textbf{32.23}                  & \textbf{59.66}                   \\ \midrule
TimeSformer \cite{bertasius2021space}                      & 2021          & Token               & Transformer       & 44.25                   & 74.03                   \\
Swin-T \cite{liu2022video}               & 2022          & Token               & Transformer       & \textbf{48.06}          & \textbf{74.47}          \\
ESTF \cite{wang2022hardvs}               & 2022          & Token               & ResNet18          & 24.68                   & 50.18                   \\
GET \cite{peng2023get}                   & 2023          & Token               & Transformer       & 37.28                   & 61.59                   \\ \midrule
Spikformer \cite{zhou2022spikformer}     & 2022          & Spike               & Transformer       & \textbf{36.94}                   & \textbf{62.37}                   \\
SDT \cite{yao2024spike}                  & 2024          & Spike               & Transformer       & 35.43                   & 58.81                    \\ 
\bottomrule   
\end{tabular}
}
\end{table}

\subsubsection{Evaluation of different groups}

We conducted fine-grained group testing on models trained on DailyDVS-200 dataset, including Slowfast\cite{feichtenhofer2019slowfast}, TSM\cite{lin2019tsm}, EST\cite{gehrig2019end}, Swin-T\cite{liu2022video}, Spikformer\cite{zhou2022spikformer}, and the results are presented in \Cref{tab:group}.

We observed that within the same framework, the performance varied significantly depending on factors such as different camera motion states, action scopes, number of actors, and camera heights. For instance, Swin-T\cite{liu2022video} exhibited different performance under CM (27.84\% vs 58.05\%), AR (59.24\% vs 43.15\% vs 34.59\%), PN (44.7\% vs 66.88\%), and HE (44.46\% vs 52.14\%). Notably, the impact of lighting conditions on model performance differed from traditional video classification. We found that the recognition performance was optimal under side lighting, followed by backlit conditions, while it was poorest under front-facing lighting.

In the testing of the same attribute category, such as camera motion, we observed that token-based models (Swin-T\cite{liu2022video}) outperformed Learnable-based model EST\cite{gehrig2019end}, frame-based traditional models (Slowfast\cite{feichtenhofer2019slowfast} and TSM\cite{lin2019tsm}) and spiking-based models (Spikformer\cite{zhou2022spikformer} and SDT\cite{yao2024spike}) (27.84\% vs 13.52\% vs 26.88\% vs 21.05\% vs 13.74\%). We analyzed this phenomenon, attributing it to the presence of excessive background events in mobile camera settings. Event-based recognition models, due to their high frame rate nature, tend to focus more on background noise, thereby impacting the model's recognition capability. Conversely, token-based models mitigate this effect. As shown in \cref{fig:data_static} (d), the number of events in mobile camera settings is significantly higher than in stationary camera settings.

% \begin{wrapfigure}{r}{5cm}
% % \vspace{-0.3cm} 1
% \centering
% \includegraphics[width=0.4\textwidth]{figure/rate3.png}
% \caption{Evaluation of using different sizes of Moving camera set for action recognition.}
% \label{fig:rate}
% % \vspace{-0.3cm} 1
% \end{wrapfigure}

\begin{table}[tb]
\caption{Fine-grained group testing on different models. Best and second best results are \textbf{highlighted} and \underline{underlined}.
}
\tabcolsep=2pt
% \small
\centering
\label{tab:group}
\resizebox{1\textwidth}{!}{
\begin{tabular}{l|cc|cc|ccc|cc|cc|cc|cc}
\toprule
\multirow{2}{*}{\textbf{Methods}} & \multicolumn{2}{c|}{\textbf{1. BC}} & \multicolumn{2}{c|}{\textbf{2. DR}} & \multicolumn{3}{c|}{\textbf{3. IL}}                  & \multicolumn{2}{c|}{\textbf{4. DT}} & \multicolumn{2}{c|}{\textbf{5. HE}} & \multicolumn{2}{c|}{\textbf{6. LO}} & \multicolumn{2}{c}{\textbf{7. CM}} \\ \cmidrule{2-16} 
                                                       & \textbf{\scalebox{1}{Complex}}  & \textbf{Easy}   & \textbf{Long}    & \textbf{Short}   & \textbf{Back} & \textbf{Front} & \textbf{Side} & \textbf{Far}     & \textbf{Near}    & \textbf{High}    & \textbf{Low}     & \textbf{Indoor}  & \textbf{Outdoor} & \textbf{Move}   & \textbf{Static}  \\ \midrule
SlowFast\cite{feichtenhofer2019slowfast}                                               & \underline{45.98}             & 39.20            & \underline{40.94}            & 42.18            & \underline{44.05}           & 34.52            & 46.13           & \underline{41.62}            & 41.64            & 38.34            & 45.36            & 43.21            & 39.08            & 21.05           & \underline{51.81}            \\
TSM\cite{lin2019tsm}                                                    & 45.37             & \underline{40.84}           & 39.68            & \underline{44.69}            & 42.91           & \underline{36.60}             & \underline{47.16}           & 40.71            & \underline{44.09}            & \underline{38.90}             & \underline{46.51}            & \underline{43.41}            & \underline{40.93}            & \underline{26.88}           & 50.16            \\
EST\cite{gehrig2019end}                                                    & 35.22             & 30.59           & 29.42            & 34.52            & 32.89           & 24.71            & 38.26           & 28.93            & 35.28            & 27.45            & 37.64            & 33.76            & 29.82            & 13.52           & 41.51            \\
Spikformer\cite{zhou2022spikformer}                                             & 38.96             & 33.45           & 33.86            & 36.68            & 36.77           & 30.09            & 39.11           & 31.57            & 39.00            & 31.82            & 39.52            & 38.54            & 30.33            & 13.74           & 46.15            \\
Swin-T\cite{liu2022video}                                                 & \textbf{52.86}    & \textbf{45.37}  & \textbf{48.57}   & \textbf{47.64}   & \textbf{51.61}  & \textbf{39.11}   & \textbf{53.39}  & \textbf{47.16}   & \textbf{48.89}   & \textbf{44.46}   & \textbf{52.14}   & \textbf{50.15}   & \textbf{44.70}    & \textbf{27.84}  & \textbf{58.05}   \\
 \bottomrule

\end{tabular}
}

% \vspace{5pt} 1

\resizebox{1\textwidth}{!}{
\begin{tabular}{l|cc|cc|cc|ccc|cc|cc|cc}
\toprule
\multirow{2}{*}{\textbf{Methods}} & \multicolumn{2}{c|}{\textbf{8. PN}} & \multicolumn{2}{c|}{\textbf{9. PE}} & \multicolumn{2}{c|}{\textbf{10. PR}} & \multicolumn{3}{c|}{\textbf{11. AR}}                 & \multicolumn{2}{c|}{\textbf{12. SH}} & \multicolumn{2}{c|}{\textbf{13. PO}} & \multicolumn{2}{c}{\textbf{14. DI}} \\ \cmidrule{2-16} 
                                       & \textbf{One}     & \textbf{Two}     & \textbf{Frontal} & \textbf{Lateral} & \textbf{No}       & \textbf{Yes}     & \textbf{Full-body} & \textbf{Limbs} & \textbf{Micro} & \textbf{No}       & \textbf{Yes}     & \textbf{Stand} & \textbf{Sit} & \textbf{Day}     & \textbf{Night}   \\ \midrule
SlowFast\cite{feichtenhofer2019slowfast}                               & 39.15            & 55.57            & 44.52            & 39.04            & 41.81             & \underline{41.48}            & 52.05              & \underline{37.59}          & 25.16          & 37.07             & \underline{42.73}            & 42.93             & 35.92            & 43.10             & \underline{33.65}            \\
TSM\cite{lin2019tsm}                                    & \underline{40.07}            & \underline{55.90}             & \underline{46.13}            & \underline{39.17}            & \underline{46.66}             & 38.98            & \underline{53.31}              & 37.21          & \underline{33.02}          & \underline{42.75}             & 42.39            & \underline{43.11}             & \underline{39.61}            & \underline{44.20}             & 33.02            \\
EST\cite{gehrig2019end}                                    & 29.88            & 45.56            & 33.42            & 31.15            & 34.81             & 30.08            & 41.18              & 28.55          & 19.81          & 35.18             & 31.55            & 32.70              & 30.13            & 34.74            & 18.55            \\
Spikformer\cite{zhou2022spikformer}                             & 33.88            & 44.10            & 37.09            & 33.94            & 37.39             & 33.80            & 43.86              & 31.95          & 23.27          & 39.47             & 34.45            & 35.01             & 37.24            & 38.30            & 19.81            \\
Swin-T\cite{liu2022video}                                 & \textbf{44.70}    & \textbf{66.88}   & \textbf{50.98}   & \textbf{45.43}   & \textbf{51.35}    & \textbf{45.33}   & \textbf{59.24}     & \textbf{43.15} & \textbf{34.59} & \textbf{46.78}    & \textbf{48.36}   & \textbf{48.96}    & \textbf{44.08}   & \textbf{50.22}   & \textbf{36.32}   \\
 \bottomrule
\end{tabular}
}
% \vspace{-0.4cm} 1
\end{table}

\begin{table}[tb]
\caption{Evaluation of existing models with different frame settings on our dataset. Best and second best results are \textbf{highlighted} and \underline{underlined}.
}
\tabcolsep=5pt
\label{tab:frame}
\centering
\resizebox{1\textwidth}{!}{
\begin{tabular}{lccc|ccc|ccc}
\toprule
\multicolumn{1}{c}{\multirow{2}{*}{\textbf{Methods}}} & \multicolumn{3}{c|}{\textbf{0.5s}}                                                        & \multicolumn{3}{c|}{\textbf{0.25s}}                                                      & \multicolumn{3}{c}{\textbf{0.125s}}                                                      \\ \cmidrule{2-10} 
\multicolumn{1}{c}{}                        & \multicolumn{1}{c}{gap0} & \multicolumn{1}{c}{gap2}  & \multicolumn{1}{c|}{gap4} & \multicolumn{1}{c}{gap0} & \multicolumn{1}{c}{gap2} & \multicolumn{1}{c|}{gap4} & \multicolumn{1}{c}{gap0} & \multicolumn{1}{c}{gap2}  & \multicolumn{1}{c}{gap4} \\ \midrule
C3D\cite{tran2015learning}                                 & 21.99                    & \multicolumn{1}{c}{14.98} & 11.70                     & 31.10                     & 24.82                    & 22.82                    & 44.81                    & \multicolumn{1}{c}{32.62} & 27.75                    \\
I3D \cite{carreira2017quo}                               & 32.30                     & 22.94                     & 20.82                    & 45.39                    & 29.54                    & 29.42                    & 59.10                     & 36.70                      & 37.50                     \\
R2Plus1D\cite{tran2018closer}                & 36.06                    & \underline{26.39}                     & \textbf{24.97}                    & \underline{49.65}                    & \underline{36.62}                    & \underline{32.10}                     & 58.88                    & \underline{48.06}                     & 40.29                    \\
SlowFast\cite{feichtenhofer2019slowfast}                           & \textbf{41.49}                    & \textbf{26.90}                     & \underline{24.55}                    & \textbf{52.16}                    & 33.28                    & 25.43                    & \textbf{64.09}                    & 44.81                     & \underline{44.64}                    \\
TSM \cite{lin2019tsm}                               & \underline{40.87}                    & 23.67                     & 22.97                    & 49.55                    & \textbf{37.94}                    & \textbf{32.37}                    & \underline{61.76}                    & \textbf{51.48}                     & \textbf{48.77}                    \\ \bottomrule
\end{tabular}
}
% \vspace{-0.4cm} 1
\end{table}

\begin{figure}[tb]
  \centering
  \includegraphics[height=4.8cm]{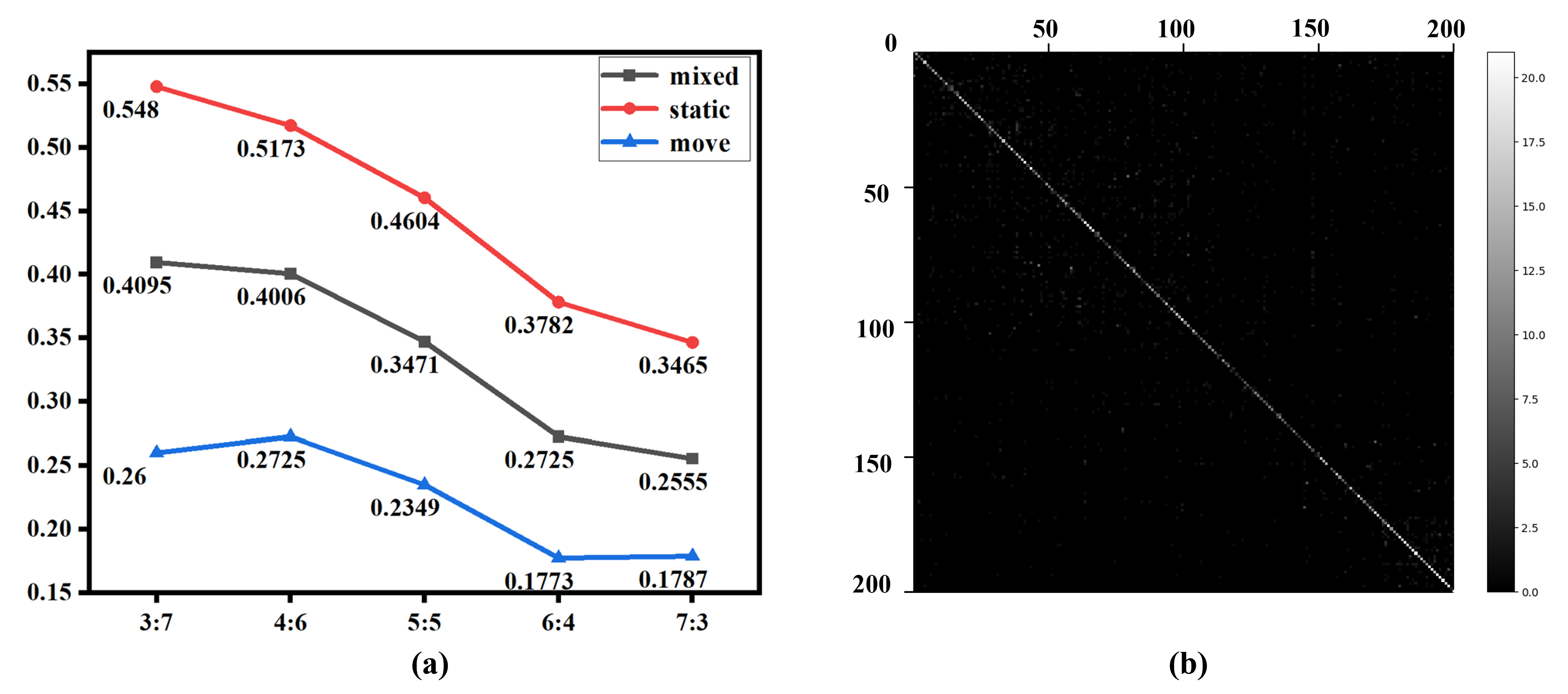}
  \caption{(a) Evaluation of using different sizes of Moving camera set for action recognition. (b) Confusion matrix of Swin-T\cite{liu2022video}.}
  \label{fig:final}
  % \vspace{-.4cm} 1
\end{figure}

To mitigate the bias in testing groups to camera motion attributes, we conducted training and testing on the TSM\cite{lin2019tsm} model using different proportions of dynamic and static data. Our experimental results, as illustrated in \cref{fig:final} (a), show a noticeable decrease in the accuracy of action evaluated under both dynamic and static cameras as the proportion of dynamic data increases. Contrary to expectations, there was no improvement in accuracy with the increasing volume of event data captured by dynamic cameras. This suggests that current action recognition models struggle to effectively learn action data captured by dynamic cameras, highlighting an urgent issue that requires resolution.
% \vspace{-0.5cm} 1
\subsubsection{Evaluation of different frame settings}

\Cref{tab:frame} provides a detailed comparison of several widely used frame-based action recognition models on DailyDVS-200 dataset with different frame settings. Since frame-based action recognition models reconstruct frame sequences from event stream data \cite{rebecq2017real}, frame sequences formed with different time step lengths often exhibit significant differences. We generated reconstructed frame sequences using three different fixed time step lengths of 0.5s, 0.25s, and 0.125s, respectively, and conducted experiments with frame intervals (gap) of 0, 2, and 4. Our experimental results revealed that increasing the event step length to capture additional temporal information contributes to improving the model's accuracy. However, increasing the frame interval leads to partial information loss, which is detrimental to model recognition.

\begin{table}[tb]

\caption{Top-10 accurate and top-10 incorrect actions of different methods. The same action is marked with the same color.}
\label{tab:t5}
\resizebox{1\textwidth}{!}{
\tabcolsep=5pt
\begin{tabular}{cll|cll}
\toprule
\textbf{Methods} & \multicolumn{1}{c}{\textbf{\makecell[c]{Top-10 \\ accurate actions}}}                                                                                                                                                                                                                                                                                  & \multicolumn{1}{c|}{\textbf{\makecell[c]{Top-10 \\ incorrect actions}}}                                                                                                                                                                                                                                          & \textbf{Methods}    & \multicolumn{1}{c}{\textbf{\makecell[c]{Top-10 \\ accurate actions}}}                                                                                                                                                                                                                                                                   & \multicolumn{1}{c}{\textbf{\makecell[c]{Top-10 \\ incorrect actions}}}                                                                                                                                                                                                                                            \\ \midrule
Slowfast\cite{feichtenhofer2019slowfast}     & \begin{tabular}[c]{@{}l@{}}1. pull out the chair \\ 2. open curtains \\ 3. \textbf{\textcolor{NavyBlue}{hand in hand circling}}  \\ 4. close the door \\ 5. wash towels  \\ 6. fall down \\ 7. turn the light on \\ 8. \textbf{\textcolor{ForestGreen}{wipe the table}} \\ 9. cross your legs  \\ 10. \textbf{\textcolor{blue}{push-up}}\end{tabular}                                                         & \begin{tabular}[c]{@{}l@{}}1. hammer table \\ 2. \textbf{\textcolor{Maroon}{play table tennis}}  \\ 3. open window  \\ 4. clean windows  \\ 5. \textbf{\textcolor{red}{pitch}}  \\ 6. charge a phone \\ 7. arrang cards  \\ 8. V sign  \\ 9. use a tablet  \\ 10. kick a ball\end{tabular}       & EST\cite{gehrig2019end}        & \begin{tabular}[c]{@{}l@{}}1. tie shoelaces \\ 2. mutual bow  \\ 3. turn the light on  \\ 4. \textbf{\textcolor{blue}{push-up}} \\ 5. turn the light off  \\ 6. fall down  \\ 7. lie on the table  \\ 8. close window  \\ 9. cheers  \\ 10. \textbf{\textcolor{NavyBlue}{hand in hand circling}}\end{tabular}   & \begin{tabular}[c]{@{}l@{}}1. \textbf{\textcolor{Maroon}{play table tennis}} \\ 2. open window \\ 3. close curtains \\ 4. slap the table  \\ 5. \textbf{\textcolor{red}{pitch}} \\ 6. OK sign  \\ 7. make paper cuttings  \\ 8. charge a phone\\ 9. hit people with things  \\ 10. headache\end{tabular} \\ \midrule
Swin-T\cite{liu2022video}  & \begin{tabular}[c]{@{}l@{}}1. \textbf{\textcolor{NavyBlue}{hand in hand circling}}  \\ 2. arm wrestling \\ 3. \textbf{\textcolor{blue}{push-up}} \\ 4. close curtains \\ 5. open curtains \\ 6. close the door  \\ 7. sit-up  \\ 8. \textbf{\textcolor{ForestGreen}{wipe the table}} \\ 9. moves heavy objects \\ 10. cross legs\end{tabular} & \begin{tabular}[c]{@{}l@{}}1. hammer table  \\ 2. \textbf{\textcolor{red}{pitch}} \\ 3. take off headphones\\ 4. write \\ 5. blow nose\\ 6. V sign \\ 7. crush paper into a ball \\ 8. take something from bag  \\ 9. chest pain \\ 10. open the bottle\end{tabular} & Spikformer\cite{zhou2022spikformer} & \begin{tabular}[c]{@{}l@{}}1. \textbf{\textcolor{ForestGreen}{wipe the table}} \\ 2. close curtains \\ 3. stand up \\ 4. \textbf{\textcolor{NavyBlue}{hand in hand circling}} \\ 5. lie on the table\\ 6. mutual bow  \\ 7. cross legs  \\ 8. clean the windows  \\ 9. \textbf{\textcolor{blue}{push-up}} \\ 10. go upstairs\end{tabular} & \begin{tabular}[c]{@{}l@{}}1. \textbf{\textcolor{red}{pitch}}\\ 2. \textbf{\textcolor{Maroon}{play table tennis}}  \\ 3. plug in the power strip\\ 4. take off shoes\\ 5. trim nails  \\ 6. stomachache  \\ 7. play with hair \\ 8. roll up sleeves \\ 9. backache  \\ 10. headache\end{tabular} \\ \bottomrule
\end{tabular}
}
% \vspace{-0.4cm} 1
\end{table}

\subsubsection{Detailed analysis to actions and methods}

We also conducted a detailed analysis of the performance of different action recognition methods on our proposed DailyDVS-200 dataset. We took four best-performing methods (SlowFast\cite{feichtenhofer2019slowfast}, Swin-T\cite{liu2022video}, EST\cite{gehrig2019end}, Spikformer\cite{zhou2022spikformer}) as examples for analysis. Firstly, we plot the confusion matrices of these methods. The confusion matrix of Swin-T\cite{liu2022video} is shown in \cref{fig:final} (b) as an example, we can see there are still many actions cannot be correctly classified. Specifically, considering the large number of action categories, we also analyzed the top-10 accurate actions and top-10 incorrect actions for each method (see \Cref{tab:t5}).

Among them, we found that the actions \emph{hand in hand circling} and \emph{push-up} ranked in the top-10 in accuracy across all four models. This could be attributed to uniqueness and large motion range of these two actions, resulting in strong spatiotemporal features. However, we also observed that some actions, such as \emph{pitch} and \emph{play table tennis} were prone to misclassification in most models. A possible explanation is the fast speed and short duration of these actions, which pose challenges for feature extraction, thereby affecting the accuracy of classification. Furthermore, we found that two-person actions were rarely in the top-10 incorrect actions, while micro-actions such as the \emph{V sign} and \emph{OK sign} were seen difficult to accurately recognize. Another notable phenomenon is that \emph{closing curtains} was well recognized in Swin-T\cite{liu2022video}, however, poorly recognized in EST\cite{gehrig2019end}, indicating significant variations among different models. Additionally, we conducted detailed testing in different scenarios and found that performance was consistently poor across all models in scenes labeled as \emph{with shadow}, \emph{front light}, \emph{long distance}, \emph{high-angle shot}, \emph{night} and \emph{outdoor}, which needed to pay more attention in the future. You can see the supplementary material for additional scenes evaluation details.

\section{Conclusion}
In this study, we introduce a comprehensive and challenging large-scale event-based action recognition dataset, named DailyDVS-200, which serves as a new benchmark for event-based action recognition. This dataset consists of 200 categories of everyday human actions performed by 47 individuals, resulting in over 22,000 event sequences. We meticulously consider the complexity of scenes, diversity of subjects, and variability of actions, providing 14 different attribute annotations for each data sample. Through evaluations and intragroup testing of over 10 different methods, we gain a nuanced understanding of the various scenarios in real-life settings where event cameras are utilized. Furthermore, comparison with other large-scale event datasets reveals that existing datasets exhibit high performance with traditional models, thus hindering innovation in event-based methods and the full utilization of event cameras' advantages. By introducing the proposed DailyDVS-200 dataset, we aim to provide new research directions for methodological innovation in this field.
% \clearpage  % TODO FINAL: This \clearpage needs to be removed from both review and camera-ready versions.

\section*{Acknowledgements}
This work was partly supported by the Chinese Defense Advance Research
Program (50912020105).

\bibliographystyle{splncs04}
\bibliography{main}
\end{document}

\makeatletter
\def\@fnsymbol#1{%
   \ifcase#1\or
   %\TextOrMath\textasteriskcentered *\or
   \TextOrMath \Letter \dagger\or
   \TextOrMath \textdagger \ddagger\else
   \@ctrerr \fi
}
\makeatother

\title{DailyDVS-200: A Comprehensive Benchmark Dataset for Event-Based Action Recognition \\ (Supplementary Material)} 

\titlerunning{DailyDVS-200 for Event-Based Action Recognition}

\author{Qi Wang\inst{1*}\orcidlink{0009-0004-4576-4458}\and
Zhou Xu\inst{1*}\orcidlink{0009-0004-5499-0537} \and
Yuming Lin\inst{1}\orcidlink{0009-0009-7439-7462} \and 
Jingtao Ye\inst{1}\orcidlink{0009-0002-6636-5072} \and
Hongsheng Li\inst{1}\orcidlink{0000-0002-9929-4023} \and
Guangming Zhu\inst{1}\orcidlink{0000-0003-3214-4095} \and
Syed Afaq Ali Shah\inst{2}\orcidlink{0000-0003-2181-8445} \and
Mohammed Bennamoun\inst{3}\orcidlink{0000-0002-6603-3257} \and
Liang Zhang\inst{1}\thanks{ Corresponding author. * Equal contribution.}\orcidlink{0000-0003-4331-5830}
}

\authorrunning{Q.~Wang et al.}

\institute{Xidian University, School of Computer Science and Technology, China \email{\{qiwang0720, xuzhou0112\}@stu.xidian.edu.cn, liangzhang@xidian.edu.cn} \and
Edith Cowan University \email{afaq.shah@ecu.edu.au} \and
University of Western Australia \email{mohammed.bennamoun@uwa.edu.au}
}

\maketitle

\begin{figure}[htb]
  \centering
  \includegraphics[height=6.9cm]{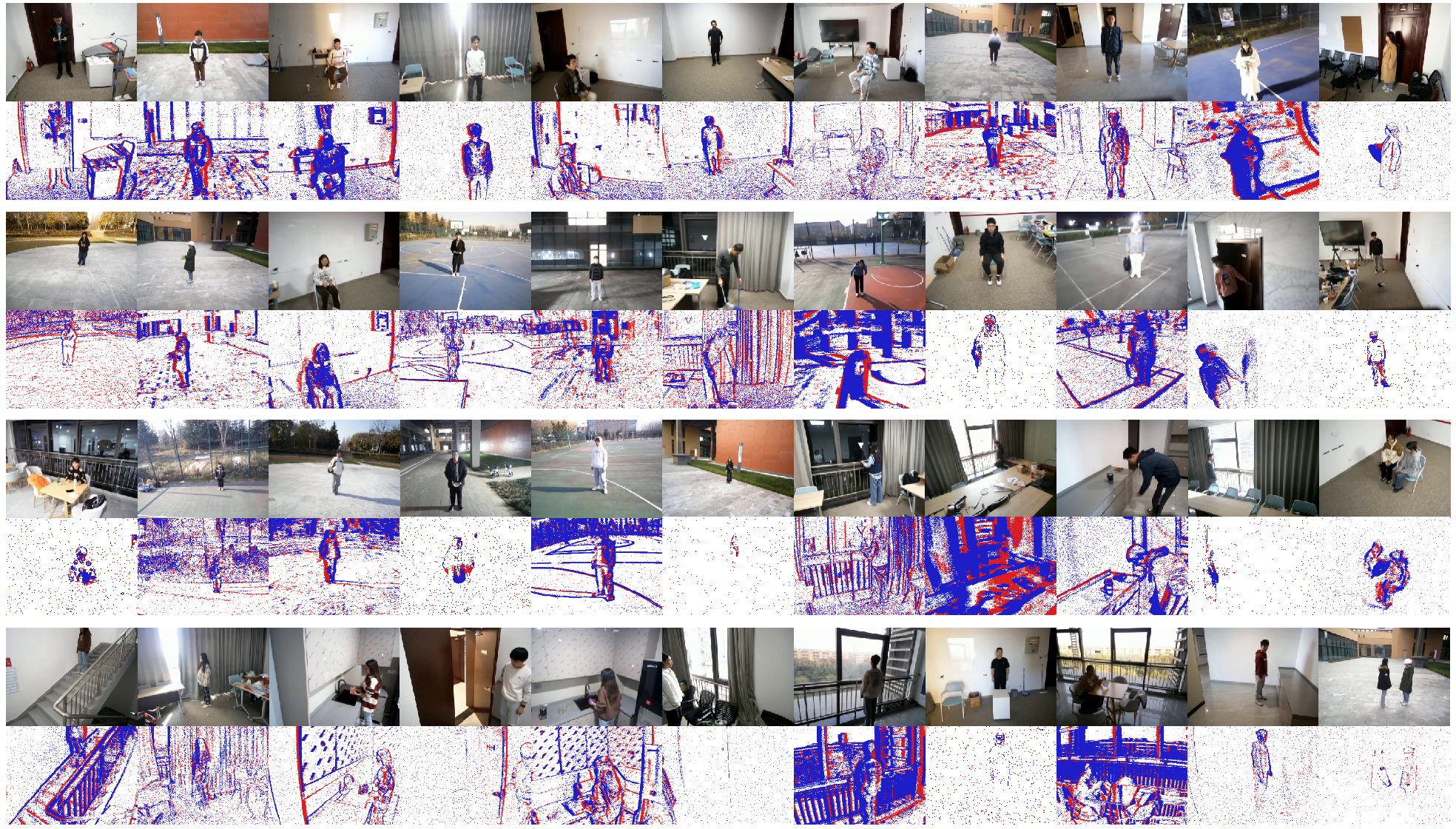}
  \caption{Example of our DailyDVS-200 dataset, including most individuals and scenes.
  }
  \label{fig:overview}
% \vspace{-.4cm}
\end{figure}

This supplementary material provides more details on our DailyDVS-200 dataset. We first introduced the detailed annotation method of the dataset and provided an example of annotation. Then, we presented specific test results under different scenarios and supplemented the analysis of factors influencing the scenarios. Moreover, we provided a detailed list of actions under our dataset, as shown in \Cref{tab:action}. Finally, we validate the challenge of our dataset by applying the same experimental settings to an existing large-scale action recognition dataset. And as shown in \cref{fig:overview}, we also provided examples of most of the scenes and individuals.

\begin{figure}[tb]
  \centering
  \includegraphics[height=3.5cm]{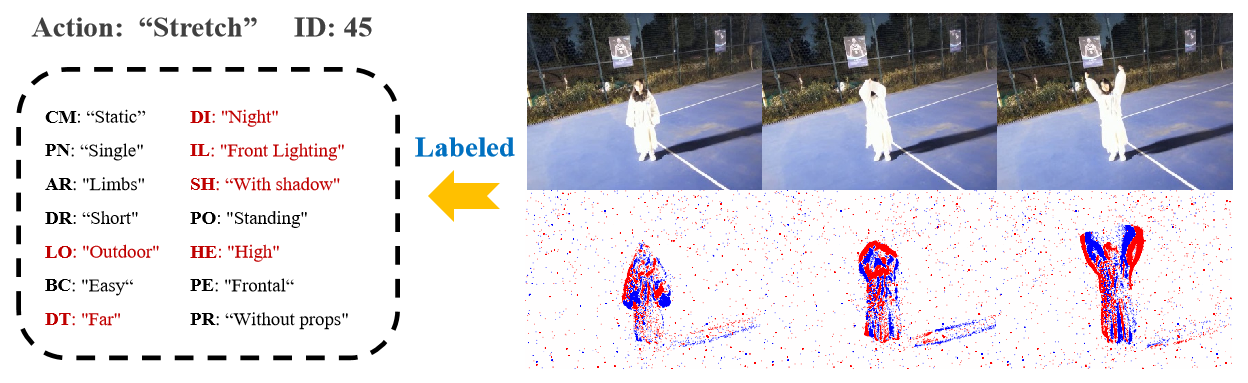}
  \caption{Examples of our attribute annotations.
  }
  \label{fig:label}
% \vspace{-.4cm}
\end{figure}

\begin{table}[htb]
        \caption{The standard for grouping and labeling attributes within our dataset.
        }
        \small
        % \vspace{-3mm}
        \tabcolsep=4pt
        \label{tab:attribute}
	\centering
    \resizebox{1\textwidth}{!}{
	\begin{tabular}{llll} \\ \toprule
    \textbf{ID} &\textbf{Attribution} &  \textbf{Standards for annotation} \\ \midrule
    PR &Props &  Whether to interact with the prop \\
    PO &Posture &  Stand or not \\
    DR & Duration & Whether the action continues to change \\
    AR &Action Range &  Body swing amplitude \\
    PN & PersonNum &  The number of subject \\
    IL & Illumination Direction & The direction of illumination relative to the person \\
    BC &Background Complexity &  Purity of background \\
    PE &Perspective &  Camera Angle \\
    DI &Diurnality &  Is it day or night \\
    LO &Location &  Is the shooting location indoors or outdoors \\
    DT &Distance &  Shooting distance greater than 5 meters is far, otherwise close \\
    HE &Height &  Shooting height greater than 2 meters is high, otherwise low \\
    SH &Shadow &  Whether the subject's shadow is captured \\
    CM &Carmera Motion &  Whether the camera moves \\
    \bottomrule
	\end{tabular}
    }
    % \vspace{-0.3cm}
\end{table}

\section{Data Annotation}
We have annotated each data with 14 attributes, as shown in the first two columns of \Cref{tab:attribute}. To provide the community with a deeper understanding of our data, we present the standards for each attribute. As shown in \Cref{tab:attribute}, we retained synchronized RGB streams during shooting, allowing us to accurately annotate the data directly through the RGB streams. Among the attributes, some can be directly judged through the RGB streams: \textbf{Props}, \textbf{Posture}, \textbf{PersonNum}, \textbf{Location}, \textbf{Diurnality}, and \textbf{Camera Motion}. For example, if there is only one person in the image, it is labeled as "Single", and if there are two people, it is labeled as "Double". 

Regarding \textbf{Duration}, we classify actions into short and long based on their duration, with actions lasting less than 5 seconds considered short and those lasting more than 5 seconds considered long. 

For \textbf{Action Range}, we divide actions into three parts based on the range of motion: full-body actions, limb actions, and micro actions. Full-body actions involve the entire body, limb actions involve only a large part of the body such as the arms or legs, and micro actions mainly involve small movements such as finger movements.

For \textbf{Illumination Direction}, due to the complexity of lighting conditions in actual images, we infer the direction of light based on the direction of the most prominent shadow of the person. 

For \textbf{Background Bomplexity}, we do not have quantitative criteria but directly classify it based on the number of contours in the background image of the scene. 

For \textbf{Perspective}, if the vertical plane facing the camera is set as 0 degrees, we classify data from -45 degrees to 45 degrees as frontal view and data from 45 degrees to 135 degrees as lateral view. 

For \textbf{Distance}, we classify data where the distance from the camera to the person is less than 5 meters as near and data where it is more than 5 meters as far. 

For \textbf{Height}, we classify data where the camera height is greater than 2 meters as high and data where it is less than 2 meters as low. 

For \textbf{shadow}, we classify based on whether there are obvious shadows in the event flow image of the data.

\section{Scene Analysis}
% \vspace{-0.5cm}

Due to the variations in our scenes under different individuals, to further demonstrate our analysis of action recognition performance in different scenes, we conducted separate tests using Swin-T\cite{liu2022video} on individuals in the validation and test sets. Specific details are shown in \Cref{tab:testing}. We sorted the accuracy of different IDs in descending order and listed the scene annotations for each ID. We found that the recognition accuracy was lowest in the scenes under ID 45, at only 0.2063. Analyzing the scene attributes, we observed that the scene was outdoors, at night, with shadows, standing, and shot from a high and far distance. We believe that this might be due to the higher contrast between the person and the background in outdoor nighttime scenes, leading to more noise due to artificial light sources and camera position being higher and further away, resulting in poorer performance in action recognition.

Furthermore, we also observed that the accuracy of scene testing at night was consistently lower than during the daytime, which aligns with traditional RGB cameras, indicating that nighttime remains a challenging scenario. Additionally, we found that scenes where individuals are sitting generally have higher accuracy compared to scenes where they are standing. One possible reason for this could be that when a person is sitting, their movements are clearer and have more complete features.

Moreover, we discovered that performance was relatively poor when the shooting position was higher and closer. We speculate that this could be due to the overhead perspective causing some degree of occlusion in movements, resulting in information loss and lower overall recognition accuracy.
% \usepackage{multirow}
\begin{table}[htb]
        \caption{Testing the individuals in the validation and test datasets within various scenes. ID represents the assigned sequence number for each individual.}
        \small
        % \vspace{-3mm}
        \tabcolsep=4pt
        \centering
        \label{tab:testing}
\resizebox{0.95\textwidth}{!}{
\begin{tabular}{lccccccc}
\toprule
\multirow{2}{*}{\textbf{ID}} & \multirow{2}{*}{\textbf{Acc}} & \multicolumn{6}{c}{\textbf{Scenes attributes}}                                                                    \\ \cmidrule{3-8} 
                             &                               & Location & Diurnality & Shadow & Posture & Height & Distance \\ \midrule
31                           & 0.5162                        & Outdoor           & Day               & Yes             & Standing          & Low             & Far               \\
4                            & 0.5054                        & Indoor            & Day               & Yes             & Standing          & Low             & Far               \\
7                            & 0.4645                        & Indoor            & Day               & Yes             & Standing          & Low             & Near              \\
3                            & 0.4577                        & Indoor            & Day               & No              & Standing          & Low             & Near              \\
42                           & 0.4457                        & Outdoor           & Day               & Yes             & Standing          & High            & Far               \\
43                           & 0.4435                        & Outdoor           & Day               & Yes             & Standing          & High            & Far               \\
37                           & 0.4423                        & Outdoor           & Day               & No              & Standing          & Low             & Near              \\
16                           & 0.4419                        & Indoor            & Day               & Yes             & Standing          & High            & Far               \\
41                           & 0.4412                        & Outdoor           & \textbf{\textcolor{NavyBlue}{Night}}             & Yes             & Standing          & Low             & Near              \\
5                            & 0.4295                        & Indoor            & Day               & Yes             & Standing          & Low             & Near              \\
24                           & 0.4221                        & Indoor            & Day               & No              & \textbf{\textcolor{Maroon}{Sitting}}           & Low             & Near              \\
33                           & 0.4095                        & Indoor            & \textbf{\textcolor{NavyBlue}{Night}}             & Yes             & Standing          & Low             & Near              \\
27                           & 0.3581                        & Outdoor           & \textbf{\textcolor{NavyBlue}{Night}}             & Yes             & Standing          & Low             & Near              \\
11                           & 0.3206                        & Indoor            & Day               & Yes             & \textbf{\textcolor{Maroon}{Sitting}}           & \textbf{\textcolor{ForestGreen}{High}}            & \textbf{\textcolor{ForestGreen}{Near}}              \\
10                           & 0.3133                        & Indoor            & Day               & No              & \textbf{\textcolor{Maroon}{Sitting}}           & \textbf{\textcolor{ForestGreen}{High}}            & \textbf{\textcolor{ForestGreen}{Near}}              \\
\textbf{45}    & \textbf{0.2063}  & \textbf{Outdoor}  & \textbf{\textcolor{NavyBlue}{Night}}   & \textbf{Yes}     & \textbf{Standing}   & \textbf{High}  & \textbf{Far}               \\ \bottomrule
\end{tabular}
}
\end{table}

\begin{table}[tb]
        \caption{Comparison of our proposed dataset with other large-scale datasets. $^{\dag}$ represent the results pretrained on Kinetics-400\cite{kay2017kinetics}.
        }
        \small
        % \vspace{-3mm} 1
        \tabcolsep=10pt
        \label{tab:large}
	\centering
    \renewcommand\arraystretch{1}
    \resizebox{.6\textwidth}{!}{
	\begin{tabular}{lccc} \\ \toprule
	\textbf{Methods}         & \textbf{TSM}\cite{lin2019tsm}  & \textbf{ESTF}\cite{wang2022hardvs}  \\ \midrule
        \( \mathrm{THU}^\mathrm{E\text{-}ACT} \)-50\cite{gao2023action}  & 95.60 / $98.75^{\dag}$ & 95.25  \\
        \( \mathrm{THU}^\mathrm{E\text{-}ACT} \)-50-CHL\cite{gao2023action}  & 49.07 / $83.83^{\dag}$ & 49.50  \\
        Hardvs\cite{wang2022hardvs}          & 97.33 / $98.55^{\dag}$ & 96.67 \\
        Bullying10K\cite{dong2024bullying10k}     & 74.22 / $91.90^{\dag}$  & 84.72 \\ \midrule
        DailyDVS-200(\textbf{Ours})   & \textbf{36.05} / $\textbf{65.90}^{\dag}$  & \textbf{31.29} \\ \bottomrule
	\end{tabular}
    }
    % \vspace{-0.3cm} 1
\end{table}

\begin{table}[tb]
        \caption{A detailed list of 200 actions in our dataset.
        }
        \small
        \tabcolsep=4pt
        \label{tab:action}
	\centering
    \renewcommand\arraystretch{1}
    \resizebox{1\textwidth}{!}{
	\begin{tabular}{llll} \\ \toprule
    0: yawn & 1: drink water &
    2: V sign & 3: roll up sleeves \\ \addlinespace
    4: put hands in pockets & 5: stomachache &
    6: put on headphones & 7: put on hat \\ \addlinespace
    8: take something from bag & 9: tear thick/thin paper &
    10: smell & 11: flip a coin \\ \addlinespace
    12: cough & 13: wave hands &
    14: point to something & 15: cross arms \\ \addlinespace
    16: putting one's hands together & 17: make the quiet sign &
    18: play with hair & 19: snap fingers \\ \addlinespace
    20: cross two arms & 21: surrender &
    22: headache & 23: chest pain \\ \addlinespace
    24: backache & 25: have a neck pain &
    26: put on glasses & 27: take off the glasses \\ \addlinespace
    28: blow nose & 29: turn over the book &
    30: wipe face & 31: use mobile phone \\ \addlinespace
    32: use a tablet & 33: take off hat &
    34: wear gloves & 35: take off gloves \\ \addlinespace
    36: call up & 37: throw things &
    38: fan self & 39: wipe away tears \\ \addlinespace
    40: clean glasses & 41: balloon blowing &
    42: shining shoes & 43: put on telephone headset \\ \addlinespace
    44: take off telephone headset & 45: put on wired headset &
    46: take off wired headset & 47: selfie \\ \addlinespace
    48: open the bottle & 49: hit people with things &
    50: applaud & 51: nod \\ \addlinespace
    52: shake head & 53: sneeze &
    54: stretch & 55: rub hands \\ \addlinespace
    56: salute & 57: thumbs up &
    58: thumbs down & 59: OK sign \\ \addlinespace
    60: roll up sleeves & 61: take off headphones &
    62: put something in bag & 63: pull a napkin \\ \addlinespace
    64: hit someone & 65: give a pat on the back &
    66: clap the hands & 67: fist bump \\ \addlinespace
    68: drop bag & 69: throw hats in the air &
    70: throw rubbish & 71: shake hands \\ \addlinespace
    72: exchange things & 73: push someone &
    74: conversation between two people & 75: take pictures of others \\ \addlinespace
    76: whisper & 77: finger guessing game &
    78: put on a coat & 79: take off the coat \\ \addlinespace
    80: tie shoelaces & 81: massaging legs &
    82: vomit & 83: massaging back \\ \addlinespace
    84: rotate arm & 85: take something from pocket &
    86: crumple the paper into a ball & 87: take off shoes \\ \addlinespace
    88: brush down & 89: comb hair &
    90: check time & 91: wear a sign around your neck \\ \addlinespace
    92: apply hand cream & 93: sit on the ground &
    94: folding clothes & 95: apply face cream \\ \addlinespace
    96: wear a watch & 97: bow &
    98: open umbrella & 99: squat \\ \addlinespace
    100: pick up things & 101: mutual bow &
    102: close umbrella & 103: kick someone \\ \addlinespace
    104: two people walking hand in hand & 105: Trim nails &
    106: move a heavy object & 107: walk \\ \addlinespace
    108: walk with a backpack & 109: fall down &
    110: spinning & 111: run in place \\ \addlinespace
    112: side kick & 113: fight &
    114: moves heavy objects & 115: hand in hand circle \\ \addlinespace
    116: walk towards each other & 117: walk away from each other&
    118: follow & 119: support sb. with one's hand \\ \addlinespace
    120: hand in hand circling & 121: jump on the spot &
    122: touch nose & 123: blow out candles \\ \addlinespace
    124: one-foot jump & 125: kick &
    126: taichi & 127: punch \\ \addlinespace
    128: waddle & 129: knock down &
    130: Light candles & 131: embrace \\ \addlinespace
    132: make paper cuttings & 133: eat meal&
    134: pen spinning & 135: shuffling cards \\ \addlinespace
    136: type on a keyboard & 137: bookbinding &
    138: wear shoes & 139: rock chair \\ \addlinespace
    140: move mouse & 141: wipe the table &
    142: flip open the laptop & 143: close the laptop \\ \addlinespace
    144: push chair & 145: sit-up &
    146: swing a badminton racket & 147: throwing ball \\ \addlinespace
    148: sit down & 149: run &
    150: long jump & 151: cheers \\ \addlinespace
    152: pour water & 153: sweep the floor &
    154: mop & 155: stir \\ \addlinespace
    156: push-up & 157: brush teeth &
    158: open the door & 159: close the door \\ \addlinespace
    160: plug in the power strip & 161: pull out the chair &
    162: charge a phone & 163: knock at a door \\ \addlinespace
    164: pull the drawer & 165: stand up &
    166: cross legs & 167: lie on the table \\ \addlinespace
    168: slap the table & 169: hammer table &
    170: turn on the tap & 171: bounce volleyball on hand \\ \addlinespace
    172: open curtains & 173: close curtains &
    174: arm wrestling & 175: washing face \\ \addlinespace
    176: bounce ball & 177: walk with an umbrella &
    178: dribbling basketball & 179: kick a ball \\ \addlinespace
    180: clean the windows & 181: wash hands&
    182: open window & 183: turn the tap off \\ \addlinespace
    184: fold paper & 185: close window &
    186: jump rope & 187: pitch \\ \addlinespace
    188: lie down & 189: wash towels &
    190: cycling & 191: turn the light on \\ \addlinespace
    192: turn the light off & 193: write &
    194: play table tennis & 195: catch watch \\ \addlinespace
    196: ride an electric bike & 197: go upstairs &
    198: go downstairs & 199: arrange cards \\ \addlinespace
    \hline
	\end{tabular}
    }
    % \vspace{-0.3cm}
\end{table}

\section{Comparison with other large-scale datasets}
To demonstrate the diversity and challenging nature of our proposed dataset, we retrained and tested some of the popular models on the current large-scale event datasets \cite{gao2023action, wang2022hardvs, dong2024bullying10k} under the same training configuration. In this section, we used the framed-based model TSM\cite{lin2019tsm} and the recently proposed baseline method ESTF in Hardvs\cite{wang2022hardvs}. Firstly, we converted all the datasets into frame sequences with a time interval of 0.5s and conducted training from scratch. Due to the absence of subject partitions in some datasets, we randomly split the datasets in a 6:1:3 ratio for experimentation. Our experimental results are shown in \Cref{tab:large}. It can be noted that these models has achieved the lowest accuracy on our dataset, with a top-1 accuracy of 31.29\% only, indicating that our proposed DailyDVS-200 is currently the most challenging large-scale dataset. Meanwhile, we further evaluated TSM \cite{lin2019tsm} with Kinetics-400\cite{kay2017kinetics} pretraining and found that these models still performed effectively under the event-based frame sequence, with DailyDVS-200 remaining the most challenging dataset, achieving a top-1 accuracy of 65.9\%.

\bibliographystyle{splncs04}
\bibliography{main}